\documentclass[10pt,twocolumn,letterpaper]{article}

\usepackage[pagenumbers]{cvpr} 

\makeatletter
\@namedef{ver@everyshi.sty}{}
\makeatother

\usepackage{graphicx}
\usepackage{amsmath}
\usepackage{amssymb}
\usepackage{booktabs}
\usepackage{multirow}
\usepackage{xcolor}
\usepackage{algorithmic}
\usepackage{algorithm}
\usepackage{multicol}

\usepackage{twoopt}
\usepackage{tikz}
\usetikzlibrary{fit}
\newcommand\tikzmark[1]{%
  \tikz[remember picture,overlay]\node[inner xsep=0pt] (#1) {};}

\newcommandtwoopt\TextboxReservoir[5][2.5cm][2cm]{%
\begin{tikzpicture}[remember picture,overlay]
  \coordinate (aux) at ([xshift=#1]#4);
  \node[inner ysep=5pt,yshift=0.6ex,draw=black,thick,
    fit=(#3) (aux),baseline] 
    (box) {};
\end{tikzpicture}%
}
\newcommandtwoopt\TextboxBalancedReservoir[5][2.5cm][2cm]{%
\begin{tikzpicture}[remember picture,overlay]
  \coordinate (aux) at ([xshift=#1]#4);
  \node[inner ysep=17pt,yshift=0.5ex,draw=black,thick,
    fit=(#3) (aux),baseline] 
    (box) {};
\end{tikzpicture}%
}

\usepackage[pagebackref,breaklinks,colorlinks]{hyperref}

\usepackage[capitalize]{cleveref}
\crefname{section}{Sec.}{Secs.}
\Crefname{section}{Section}{Sections}
\Crefname{table}{Table}{Tables}
\crefname{table}{Tab.}{Tabs.}

\usepackage[accsupp]{axessibility}  

\newcommand{\crchange}[1]{#1}

\newcommand{\Data}{\mathcal{D}}
\newcommand{\Task}{\mathcal{T}}
\newcommand{\RR}{\mathbb{R}}
\newcommand{\logit}{o}
\newcommand{\acc}{\text{acc}}

\newcommand{\projacc}{\text{proj-acc}}

\newcommand{\cka}{\text{CKA}}
\newcommand{\CE}{\text{cross-entropy}}
\newcommand{\appendixnote}{Appendix}
\newcommand{\meanstd}[2]{#1{\scriptsize±#2}}
\definecolor{ao(english)}{rgb}{0.0, 0.5, 0.0}

\newcommand{\boost}[1]{\textcolor{ao(english)}{(+#1)}}
\newcommand{\reduce}[1]{\textcolor{ao(english)}{(-#1)}}

\newcommand{\argmax}{\operatornamewithlimits{argmax}}

\def\cvprPaperID{8341} 
\def\confName{CVPR}
\def\confYear{2023}

\begin{document}

\title{
Preserving Linear Separability in Continual Learning \\by Backward Feature Projection
}

\author{Qiao Gu\\
University of Toronto\\
{\tt\small qgu@cs.toronto.edu}
\and
Dongsub Shim\\
LG AI Research\\
{\tt\small dongsub.shim@lgresearch.ai}
\and
Florian Shkurti\\
University of Toronto\\
{\tt\small florian@cs.toronto.edu}
}
\maketitle

\begin{abstract}
    Catastrophic forgetting has been a major challenge in continual learning, where the model needs to learn new tasks with limited or no access to data from previously seen tasks. To tackle this challenge, methods based on knowledge distillation in feature space have been proposed and shown to reduce forgetting~\cite{douillard2020podnet, dhar2019learning, kang2022class}. However, most feature distillation methods directly constrain the new features to match the old ones, overlooking the need for plasticity. To achieve a better stability-plasticity trade-off, we propose Backward Feature Projection (BFP), a method for continual learning that allows the new features to change up to a learnable linear transformation of the old features. BFP preserves the linear separability of the old classes while allowing the emergence of new feature directions to accommodate new classes. BFP can be integrated with existing experience replay methods and boost performance by a significant margin. We also demonstrate that BFP helps learn a better representation space, in which linear separability is well preserved during continual learning and linear probing achieves high classification accuracy. The code can be found at \href{https://github.com/rvl-lab-utoronto/BFP}{https://github.com/rvl-lab-utoronto/BFP}. 
    \vspace{-1em}
\end{abstract}


\section{Introduction}
\label{sec:intro}

\begin{figure}[!t]
    \centering
    \includegraphics[width=\linewidth]{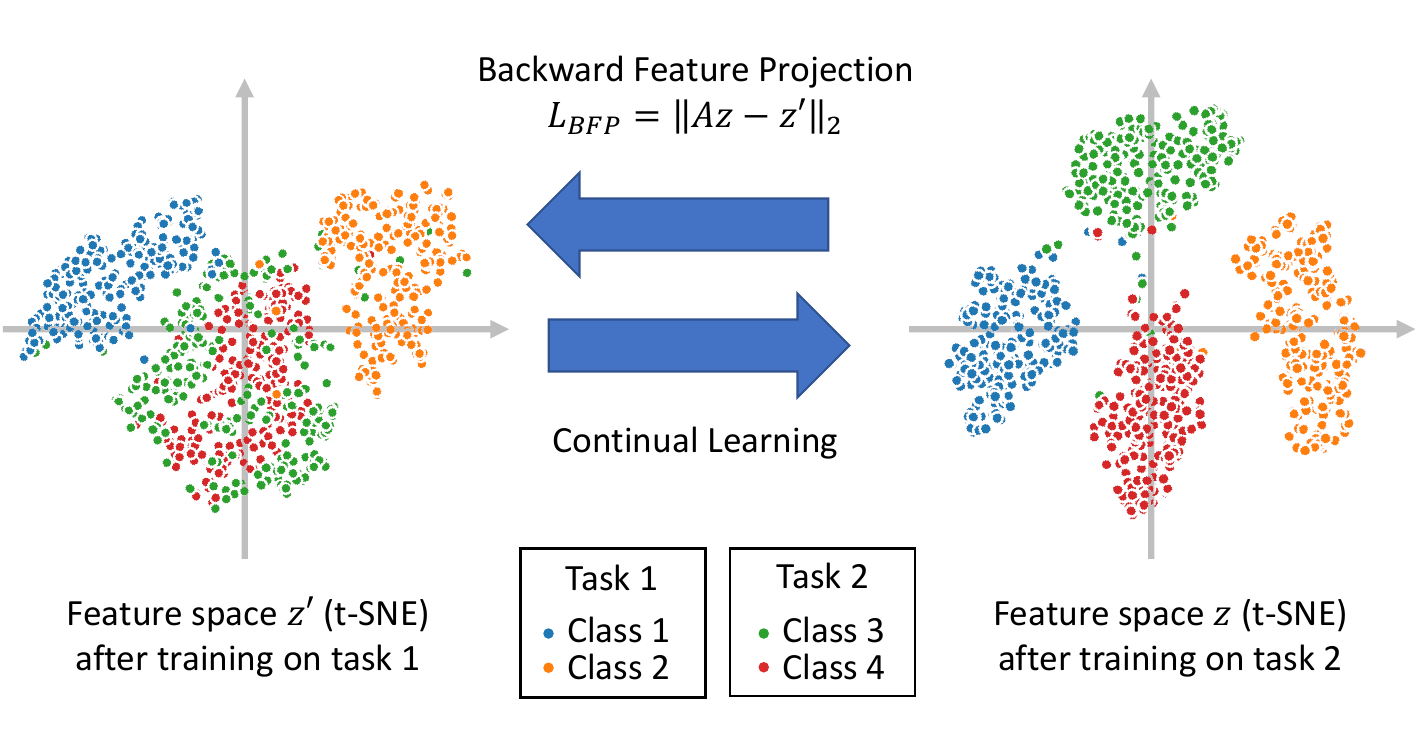}
    \caption{
    Feature distribution before and after training on a task in a class incremental learning experiment on MNIST, visualized by t-SNE. 
    \textbf{Left}: before training on task 2, seen classes (1,2) are learned to be separable along the horizontal axis for classification, while unseen classes (3, 4) are not separable. \textbf{Right}: after training on task 2, the new vertical axis is learned to separate new classes (3,4). 
    Based on this observation, we propose the Backward Feature Projection loss $L_{BFP}$, which allows new feature dimensions to emerge to separate new classes in feature space and also preserves the linear separability of old classes to reduce catastrophic forgetting. 
    }
    \label{fig:teaser}
    \vspace{-1em}
\end{figure}

Despite their many successes, deep neural networks remain prone to catastrophic forgetting~\cite{mccloskey1989catastrophic}, whereby a model's performance on old tasks degrades significantly while it is learning to solve new tasks. 
Catastrophic forgetting has become a major challenge for continual learning (CL) scenarios, where the model is trained on a sequence of tasks, with limited or no access to old training data. 
The ability to learn continually without forgetting is crucial to many real-world applications, such as computer vision~\cite{masana2022class, wang2021wanderlust}, intelligent robotics~\cite{lesort2020continual}, and natural language processing~\cite{biesialska2020continual, jang2021towards}. 
In these settings, an agent learns from a stream of new data or tasks, but training on the old data is restricted due to limitations in storage, scaling of training time, or even concerns about privacy. 

The continual learning problem has received significant attention and multiple solution themes have emerged.
Experience replay methods~\cite{buzzega2020dark, lopez2017gradient}, for example, store a limited number of (or generate) old training examples and use them together with new data in continual learning. 
Parameter regularization methods~\cite{lee2017overcoming, zenke2017continual} restrict the change of important network parameters. Knowledge distillation methods~\cite{li2017learning, dhar2019learning, douillard2020podnet} regularize the intermediate output of the CL model to preserve the knowledge from old tasks. Architectural methods~\cite{rusu2016progressive, mallya2018packnet, yan2021dynamically} adopt expansion and isolation techniques with neural networks to prevent forgetting. All these methods strive to balance learning new knowledge (plasticity) and retaining old knowledge (stability). 

We present a continual learning algorithm, focusing on knowledge distillation (KD) in feature space. 
In the continual learning context, KD treats the continual learning model as the student and its old checkpoint as the teacher and regularizes the network intermediate outputs to reduce forgetting~\cite{li2017learning, buzzega2020dark, douillard2020podnet, dhar2019learning, kang2022class, cha2021co2l, barletti2022contrastive, dhawan2023efficient}.
Although recent CL methods based on KD have been effective at reducing forgetting, they typically adopt the $L_2$ distance for distillation, forcing the learned features to be close to their exact old values. This is too restrictive and results in CL models that are more rigid in retaining old knowledge (stronger stability), but less flexible in adapting to new tasks (weaker plasticity). Our method has a better tradeoff of stability and plasticity. 

In this paper, we pay attention to the feature space in CL and study its evolution. 
We show that a small number of principal directions explain most of the variance in feature space and only these directions are important for classification. A large number of directions in the feature space have little variance and remain unused. 
When the model is trained on new tasks, new features need to be learned along those unused directions to accommodate new classes, as illustrated in Figure~\ref{fig:teaser}. Without handling forgetting, the old principal directions, along which the old classes are linearly separable, will be forgotten. 
Our results indicate that such forgetting of learned principal directions in the feature space is an important reason for catastrophic forgetting. 

Based on this insight, as shown in Figure~\ref{fig:teaser}, we propose a Backward Feature Projection (BFP) loss, an effective feature distillation loss that enforces feature consistency up to a learnable linear transformation, not imposing exact equality of features. This transformation aims to preserve the linear separability of features backward in time.
We show that this linear projection is important because it can rotate, reflect, and scale features, while maintaining the linear separability of the previously learned classes in the new feature space. 
Projecting backward allows the features to change and new decision boundaries to be learned along the unused feature directions to classify new classes. 
BFP can be integrated into existing CL methods in a straightforward way and experiments show that this simple change boosts the performance over baselines by a large margin. 

Our experiments show that the proposed BFP regularization loss can improve the baseline methods by up to 6\%-8\% on the challenging Split-CIFAR10 and Split-CIFAR100 datasets, achieving state-of-the-art class-incremental learning accuracy. More importantly, the linear probing experiments show that BFP results in a better feature space where different classes are more separable. 
See Figure~\ref{fig:teaser} for an illustrative example. 
Our contributions are as follows:
\begin{itemize}
\vspace{-0.5em}
\itemsep0em 
    \item We provide an analysis of feature space evolution during continual learning, distinguishing the important feature components from unimportant ones. 
    \vspace{-0.2em}
    \item We propose the Backward Feature Projection (BFP) loss, which preserves the linear separability of old classes while allowing plasticity during continual learning, i.e. features are allowed to change. 
    \vspace{-0.2em}
    \item When combined with simple experience replay baselines, BFP helps learn better feature space and achieves state-of-the-art performance on challenging datasets. 
\vspace{-0.4em}
\end{itemize}


\section{Related Work}
\label{sec:related}


\subsection{Experience Replay Methods}

Experience replay or rehearsal methods use a small memory buffer to keep the training data of old tasks. When the model is training on the new task, old training examples are extracted and trained together with new ones. 
Recent replay-based CL approaches mainly differ in three components, namely which examples to store, how examples are replayed, and how to update the network using old examples. 
Recent work has focused on evolving the three components mentioned above. 
ICaRL~\cite{rebuffi2017icarl} chooses examples into the memory such that the mean in the feature space of the memory buffer matches that of training data. 
MIR~\cite{aljundi2019online} prioritizes the replay of the examples that are mostly interfered with by a virtual update on the network parameters. 
DER/DER++~\cite{buzzega2020dark} augments the cross entropy loss with a logit distillation loss when the memory data is replayed. 
GEM~\cite{lopez2017gradient} and A-GEM~\cite{chaudhry2018efficient} develop optimization constraints when trained on new tasks using the old data from memory. 
Some other work~\cite{shin2017continual, liu2020generative, van2018generative} also learn to generate images for replay during CL, but the continual training of the generation network adds some additional challenges. 
Although the idea is straightforward, experience replay methods often achieve better performance than other types of methods, which marks the importance of storing the old data. 

\subsection{Parameter Regularization Methods}

Parameter regularization methods study the effect of neural network weight changes on task losses and limit the movement of important ones, which would otherwise cause forgetting on old tasks. This line of work typically does not rely on a replay buffer for old task data. 
One of the pioneering works along this line is EWC~\cite{kirkpatrick2017overcoming}, which proposed to use the empirical Fisher Information Matrix to estimate the importance of the weight and regularize the weight change in continual learning. 
SI~\cite{zenke2017continual} uses the estimated path integral in the optimization process as the regularization weight for network parameters. 
MAS~\cite{aljundi2018memory} improves this idea by adopting the gradient magnitude as a sensitivity measure. 
RWalk~\cite{chaudhry2018riemannian} combines Fisher information matrix and online path integral to approximate the parameter importance and also keeps a memory to improve results. 

\subsection{Knowledge Distillation Methods}

Originally designed to transfer the learned knowledge of a larger network (teacher) to a smaller one (student), knowledge distillation methods have been adapted to reduce activation and feature drift in continual learning. 
Different from parameter regularization methods that directly regularize the network weights, KD methods regularize the network intermediate outputs. 
Li~\etal~\cite{li2017learning} proposed an approach called Learning without Forgetting (LwF), regularizing the output logits between the online learned model and the old checkpoint. 
DER/DER++~\cite{buzzega2020dark} combines this logit regularization with experience replay and further improves the performance. 
Later Jung~\etal~\cite{jung2016less} proposed to do knowledge distillation on the feature maps from the penultimate layer and freeze the final classification layer. 
Pooled Output Distillation (PODNet)~\cite{douillard2020podnet} extended the knowledge distillation method to intermediate feature maps, and studied how different ways of feature map pooling affect continual learning performance. They proposed to pool the feature maps along the height and weight dimensions respectively to achieve a good stability-plasticity trade-off. 
Recent work~\cite{dhar2019learning, kang2022class} also used gradient information (e.g. Grad-CAM) as weighting terms in the feature distillation loss, such that feature maps that are important to old tasks will change less during continual learning. 

Different from existing KD methods, we use a learnable linear layer to project new features to the old ones. 
This idea was explored in~\cite{fini2022self, gomez2022continually}, but their work only integrated it in a contrastive learning framework and used a nonlinear mapping function. However, in this work, we use a learnable \textit{linear} transformation and formulate it as a simple knowledge distillation loss in the feature space. 
We demonstrate that our method promotes linear separability in feature space during continual learning. 
We also show the value of BFP in the supervised CL setting with experience replay, and no augmentation or contrastive learning is needed.

\begin{figure}[!t]
    \centering
    \includegraphics[width=0.48\linewidth]{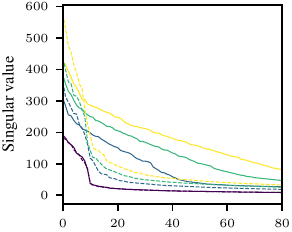}
    \includegraphics[width=0.48\linewidth]{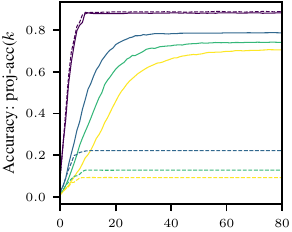}
    \includegraphics[width=0.48\linewidth]{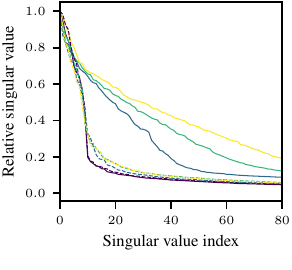}
    \includegraphics[width=0.48\linewidth]{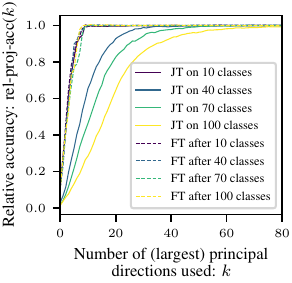}
    \caption{
    Feature distribution and contribution to classification during continual learning on Split-CIFAR100 with 10 classes per task. 
    \textbf{Left}: feature variance (singular values) along each principal direction; \textbf{right}: classification accuracies proj-acc($k$) using projected features. 
    \textbf{Upper} plots show the absolute singular values and accuracies, and \textbf{lower} ones show their relative values, normalized by the maximum on each curve. 
    Finetuning (FT) is a naive CL baseline that does not handle forgetting at all and gives a lower-bound performance. Joint Training (JT) is an oracle CL method that is jointly trained on all classes seen so far and shows the upper bound. Contrasting JT with FT reveals the ideal properties for good CL methods. 
    As the model is continually trained on more classes, more feature dimensions are learned and needed for classification. However, compared to the full feature dimension (512), only a small subspace (around 10 principal directions for 10 classes and 80 for 100 classes) is crucial for CL performance, as the relative accuracies quickly saturate with the number of principal directions used. 
    }
    \label{fig:relacc}
    \vspace{-1em}
\end{figure}

\section{Method}
\label{sec:method}


\subsection{Setting and Notation}
\label{sec:method-setting}

In a typical continual learning setting, a model $f$ is sequentially trained on a set of tasks $\Task = \{1, 2, 3, \cdots, T\}$. 
Within each task $t$, input $x$ and the corresponding ground truth output $y$ are drawn i.i.d. from the task data distribution $\Data_t=(X_t, Y_t)$ and used to train the model. Here $X_t$ and $Y_t$ denote the set of inputs and outputs from task $t$. 
To illustrate our method, the CL model is decomposed into two parts $f_\theta(x)=g_\phi(h_\psi(x))=g_\phi\circ h_\psi(x)$ with $\theta = \{\phi, \psi\}$, where $h$, parameterized by $\psi$, is a non-linear feature extractor, mapping input image $x$ to a low-dimensional feature $z \in \RR^d$. 
The classification head $g$, parameterized by $\phi$, is a linear layer that maps the latent feature $z$ to classification logits $\logit \in \RR^c$, where $c$ is the total number of classes. 
In this paper, we mainly consider the class incremental learning (Class-IL) setting and the task incremental learning (Task-IL) setting, and the proposed method works on both settings. In these settings, $\Data_t$ contains training data from a set of classes $\mathcal{C}_t$, where $\mathcal{C}_t$ are disjoint for different task $t$. 
In Task-IL, task identifiers $t$ for each input are available during evaluation time, and thus the model can focus the decision boundaries within each task. On the contrary, Class-IL requires making a decision among all classes during inference time and thus is more challenging. 
We further denote the model after training on task $j$ as $f^j = g^j \circ h^j$, and the feature extracted by $h^j$ from a datapoint in task $i$ as $z_i^j = h^j(x), \; x\in \Data_i$. 
The set of all $z_i^j$ forms a feature matrix $Z_i^j = h^j(\Data_{i}) \in \RR^{d\times n}$, and $n$ is the number of datapoints in $\Data_i$. 
And similarly, the set of features extracted from $D_1$ to $D_i$ using $h_j$ is denoted by $Z_{1:i}^j = h_j(\Data_{1:i})$. 

\subsection{Analyzing Feature Forgetting in CL}
\label{sec:method-featureforget}


Motivated by recent work showing that representation drifts in the feature space have been a major cause for catastrophic forgetting~\cite{yu2020semantic, driscoll2022representational, caccia2022erace}, we study the evolution of feature space in CL and answer two key questions: (1) how many dimensions (principal directions) in the learned feature space are occupied by the data? And (2) how many of them are used for classification? 
We answer the first question by conducting principal component analysis (PCA)~\cite{pearson1901liii} on the feature matrix $Z_{1:t}^t$, which contains feature extracted by $h^t$ from all data seen so far $\Data_{1:t}$. Suppose its singular vector decomposition gives $Z_{1:t}^t=USV^T$, and then principal directions are the left singular vectors $U=[u_1, u_2, \cdots, u_d]$, where $u_l \in \RR^d$ are sorted by the corresponding singular values $s_l$. PCA gives the data distribution of seen tasks in the feature space and thus answers the first question. 
The second question is answered by evaluating the features projected onto a subspace spanned by the first $k$ principal directions $U_{1:k}$. 
Specifically, we define the classification accuracy of the projected features as
\begin{align}
    \projacc(k)  = \acc \left( y, g^t(U_{1:k} U^T_{1:k} z ) \right)
\end{align}
where $k$ is the number of largest principal components used and \projacc~is computed over the testing set of task $t$. 
With a larger $k$, more information is retained in the projected feature $U_{1:k} U^T_{1:k} z$ and used for classification. The changes of \projacc~with the increase of $k$ reflect the importance of each principal direction being added.


In Figure~\ref{fig:relacc}, we plot $s_k$ and $\projacc(k)$ versus $k$ when a model has been trained on a certain number of classes during CL. We compare two simple CL methods: finetuning (FT) where the model is continually trained on the online data stream without any means to reduce catastrophic forgetting, and joint training (JT) where all the training data seen so far is used to train the network. Typically, FT serves as a naive lower bound for CL performance and JT an oracle upper bound. Contrasting FT and JT reveals the difference in feature space obtained from the worst and the ideal CL methods. 

We can see from Figure~\ref{fig:relacc}, for JT, as the network is trained on more classes, the learned features span a larger subspace and the classifier needs more principal directions to achieve good classification accuracies (high relative \projacc). This shows that during continual learning, more feature directions are needed to make new classes linearly separable in the feature space.
However, for the naive FT baseline, the number of principal directions with large variance does not increase with the number of seen classes.
This indicates feature forgetting: a poor CL method only focuses on the feature directions that are important for the current task. The feature directions for old tasks are suppressed to low variance and thus forgotten. 
On the other hand, compared to the full feature dimension $d=512$, JT accuracies still saturate with a relatively small $k=80$, which is roughly the number of classes seen so far. Other feature directions that have low variance are not used for classification, and such ``unused'' feature directions could leave room for future tasks in CL. 

Based on this insight, we argue for the benefit of preserving important feature directions for old tasks while allowing new ones to emerge for new tasks during continual learning. 
Therefore, we propose to learn a \textit{linear} transformation that projects the new feature space back to the old one and in the following section, we show it can achieve both goals. 


\subsection{Backward Feature Projection}
\label{sec:method-bfp}


We denote the feature extractor that is being trained on the current task $t$ as $h$, which may not have converged, and the converged model checkpoint at the end of the last task as $h'=h^{t-1}$. Given an input example $x$, the extracted features are denoted as $z = h(x)$ and $z' = h'(x)$. 
To preserve information in feature space such that the new feature $z$ should contain at least the information as that in $z'$, we can learn a projection function $p$ that satisfies $z' = p(z)$~\cite{fini2022self, gomez2022continually}.

In this work, we propose that a \textit{linear} transformation matrix $A$ can well preserve linear separability and suffice to reduce forgetting. Formally, we propose the Backward Feature Projection (BFP) loss in continual learning. Given a training example $x$,
\begin{align}
    L_{\text{BFP}} (x;\psi, A) =& \| Az - z' \|_2  \\
    =& \| A h_\psi(x) - h'(x) \|_2, \label{eq:bfp}
\end{align}
where we omit the bias term by adding a fixed entry of $1$ in the feature vector $z$. Here we only optimize $h_\psi$ and $A$, while we freeze $h'$ and thus omit its parameters. 

In the following, we show that the BFP loss can preserve the linear separability of old classes while allowing new classes to be classified along the unused directions in the old feature space.
Consider the extracted features from any two examples from the old classes $z'_i=h'(x_i),\;x_i\in C_1$ and $z'_j=h'(x_j),\; x_j\in C_2$, where $C_1, C_2 \in \mathcal{C}_{t-1}$. If they are learned to be linear separable at the end of task $t-1$, then there exists a vector $w$ and a threshold $b$, such that
\begin{align}
    w^T z'_{i} > b > w^T z'_{j},\;\forall i \in C_1,\; \forall j \in C_2. 
\end{align}
Then if the BFP loss in Equation~\ref{eq:bfp} is well optimized, i.e. $z' \approx Az$ with a linear transformation $A$. Then for the features extracted from $h$,
\begin{align}
    w^T A z_{i} >& b > w^T A z_{j},\;\forall i \in C_1,\; \forall j \in C_2 \\
    \Rightarrow (A^T w)^T z_{i} >& b > (A^T w)^T z_{j},\;\forall i \in C_1,\; \forall j \in C_2. 
\end{align}
Therefore the feature vectors $z$ from the old classes $C_1, C_2$ remain linearly separable along the direction $A^T w$ in the new feature space. The linear classifier $g$ is trained to find this decision boundary during CL with experience replay. 



To classify new classes, the network needs to map them to linearly separable regions in the feature space. 
The linear transformation in BFP achieves it by arranging the new classes along the ``unused'' feature directions that have low variance and thus are not occupied by the old tasks. 
Consider that the features extracted from future task $\Data_t$ using the old model $h'$ are probably not separable and mixed together. This is natural as $h'$ has not been trained on it. 
As we can see from Section~\ref{sec:method-featureforget} and Figure~\ref{fig:relacc}, there exists many principal directions with low variance, along which features from different classes are not separable, Ideally, a CL model should take up these ``unused'' feature directions to learn features that are needed to classify new classes. 
Without loss of generality, suppose before the model is trained on a new task $t$, the feature extracted from the new task $z' = h'(x)$, $x\in X_t$, are all mapped to zero along an ``unused'' feature direction $v$, i.e. $v^T z'= 0$. 
Then after learning on task $t$, the feature $z=h(x)$ from new classes $C_3, C_4\in \mathcal{C}_t$ can be learned to be separable along that feature direction $v$, 
\begin{align}\label{eq:vtz}
    v^T z_{i} > v^T z_{j},\;\forall i \in C_3,\; \forall j \in C_4. 
\end{align}
In this case, $A$ can be learned such that $v\notin \text{Col}(A)$ and thus $v^T (Az)=0$ while $v^T z\neq 0$ (satisfying Equation~\ref{eq:vtz}). In this way, the BFP loss in Equation~\ref{eq:bfp} allows the new class to be separable along $v$ and still can be minimized. 
Note that during the actual continual learning with BFP, neither $w$, $v$ nor the dimensionality of them is defined or needed. They are learned implicitly in the matrix $A$ through gradient descent optimization. $w$ and $v$ can be extracted and analyzed by PCA decomposition, but it is not required for training. 

\subsection{Loss functions}
\label{sec:method-loss}

We integrate the proposed backward feature projection method into an experience replay framework~\cite{buzzega2020dark}, where we keep a buffer $M$ storing training examples from old tasks. We focus on experience replay methods because they are simple and outperform other types of CL methods by a large margin according to a recent survey~\cite{masana2020class}.
We keep the model checkpoint at the end of the last task $f^{t-1}$ together with the online trained model $f$. 
During continual learning, the online model $f$ is trained on a batch from the online data stream of the current task $\Data_t$ using cross-entropy loss. 
\begin{align}\label{eq:loss-ce}
    L_{\text{ce}}(\Data_t; \theta) = \sum_{{x, y} \in  \Data_t} \CE (y, f_\theta(x))
\end{align}
Meanwhile, we sample another batch from $M$ for experience replay. Following~\cite{buzzega2020dark}, a cross-entropy loss and a logit distillation loss are applied on the replayed data
\begin{align}\label{eq:loss-repce}
    L_{\text{rep-ce}}(M; \theta)  =& \sum_{{x, y} \in M} \CE (y, f_\theta(x)), \\ \label{eq:loss-replogits}
    L_{\text{rep-logit}}(M; \theta) =& \sum_{{x, y} \in M} \| f_\theta(x) - f^{t-1}(x) \|_2^2.
\end{align}
And we apply our backward feature projection loss on both the online data stream $\Data_t$ and the replayed examples $M$
\begin{align} \label{eq:loss-BFP}
    L_{\text{BFP}}(\Data_t, M; \psi, A) =& \sum_{{x, y} \in \Data_t, M} \| A h_\psi(x) - h^{t-1}(x) \|_2. 
\end{align}
The total loss function used in continual learning is the weighted sum of the losses above. 
\begin{align}
\begin{split}\label{eq:total-loss}
    L(\Data_t, M;\theta, A) = L_{\text{ce}}(\Data_t;\theta) + \alpha L_{\text{rep-ce}}(M;\theta) \\  + \beta L_{\text{rep-logit}}(M;\theta) + \gamma L_{\text{BFP}}(\Data_t, M;\psi, A)
\end{split}
\end{align}
During training on task $t$, both the linear transformation $A$ and the model $f_\theta$ are optimized, and the old model checkpoint $f^{t-1}$ remains fixed. In our experiments, the matrix $A$ is randomly initialized at the beginning of each task. The pseudo-code of the proposed algorithm can be found in the~\appendixnote.


%

\section{Experiments}
\label{sec:exp}

\begin{table*}[t]
\centering
\footnotesize
\begin{tabular}{ll|lllll}
\multicolumn{1}{c}{Setting} & \multicolumn{1}{c|}{Method}      & \multicolumn{2}{c}{S-CIFAR10}                                                                         & \multicolumn{2}{c}{S-CIFAR100}                                                                        & S-TinyImageNet                                    \\ \hline \hline
\multicolumn{1}{r}{}        & \multicolumn{1}{c|}{Buffer Size} & 200                                               & 500                                               & 500                                               & 2000                                              & 4000                                              \\ \hline \hline
Class-IL                    & Joint Training (JT)                               & \multicolumn{2}{c}{\meanstd{91.27}{0.57}}                                          & \multicolumn{2}{c}{\meanstd{70.68}{0.57}}                                          & \meanstd{59.61}{0.25}          \\
                            & Finetuning (FT)                               & \multicolumn{2}{c}{\meanstd{36.20}{2.02}}                                          & \multicolumn{2}{c}{\meanstd{9.36}{0.07}}                                           & \meanstd{8.11}{0.08}           \\ 
                            & iCaRL~\cite{rebuffi2017icarl}   & \meanstd{63.58}{1.22}          & \meanstd{62.62}{2.07}          & \meanstd{46.66}{0.23}          & \meanstd{52.60}{0.38}          & \meanstd{31.47}{0.46}          \\
                            & FDR~\cite{benjamin2018fdr}      &\meanstd{31.24}{2.61}           &\meanstd{28.72}{2.86}           &\meanstd{22.64}{0.56}           &\meanstd{34.84}{1.03}           &\meanstd{26.52}{0.41} \\
                            & LUCIR~\cite{hou2019learning}     & \meanstd{58.53}{3.03}          & \meanstd{70.37}{0.97}          & \meanstd{35.14}{0.57}          & \meanstd{48.95}{1.21}          & \meanstd{29.79}{0.70}          \\ 
                            & BiC~\cite{wu2019large}          & \meanstd{59.53}{1.77}          & \meanstd{75.41}{1.14}          & \meanstd{35.96}{1.38}          & \meanstd{45.44}{0.96}          & \meanstd{15.98}{1.01}          \\
                            & ER-ACE~\cite{caccia2022erace}   & \meanstd{63.54}{0.42}          & \meanstd{71.17}{1.38}          & \meanstd{38.86}{0.72}          & \meanstd{50.20}{0.39}          & \meanstd{37.72}{0.16}          \\
                            \cline{2-7} 
                            & ER~\cite{riemer2019learning}      & \meanstd{58.07}{2.92}          & \meanstd{68.04}{2.18}          & \meanstd{20.34}{0.96}          & \meanstd{37.44}{1.48}          & \meanstd{23.29}{0.54}          \\
                            & ER w/ BFP (Ours)                 & \meanstd{63.27}{1.09} \boost{5.21}          & \meanstd{71.51}{1.58} \boost{3.47}          & \meanstd{22.54}{1.10}  \boost{2.20}        & \meanstd{38.92}{1.94} \boost{1.48}         & \meanstd{26.33}{0.68}  \boost{3.04}        \\ \cline{2-7} 
                            & DER++~\cite{buzzega2020dark}     & \meanstd{65.41}{1.60}          & \meanstd{72.65}{0.33}          & \meanstd{38.88}{0.91}          & \meanstd{52.74}{0.79}          & \meanstd{41.24}{0.64}          \\
                            & DER++ w/ BFP (Ours)              & \textbf{\meanstd{72.21}{0.22}} \boost{6.80} & \textbf{\meanstd{76.02}{0.79}} \boost{3.37} & \textbf{\meanstd{47.45}{1.30}} \boost{8.56} & \textbf{\meanstd{57.91}{0.66}} \boost{5.17} & \textbf{\meanstd{43.40}{0.41}} \boost{2.16} \\ \hline \hline
Task-IL                     & Joint Training (JT)                               & \multicolumn{2}{c}{\meanstd{98.19}{0.16}}                                          & \multicolumn{2}{c}{\meanstd{91.40}{0.43}}                                          & \meanstd{82.21}{0.33}          \\
                            & Finetuning (FT)                               & \multicolumn{2}{c}{\meanstd{65.01}{5.12}}                                          & \multicolumn{2}{c}{\meanstd{35.54}{2.63}}                                          & \meanstd{18.46}{1.26}          \\  
                            & iCaRL~\cite{rebuffi2017icarl}    & \meanstd{90.27}{1.21}          & \meanstd{90.05}{1.46}          & \meanstd{84.45}{0.48} & \meanstd{86.24}{0.47}          & \meanstd{66.06}{0.75}          \\
                            & FDR~\cite{benjamin2018fdr}      &\meanstd{91.42}{1.03}           &\meanstd{93.40}{0.31}           &\meanstd{74.66}{0.16}         &\meanstd{82.15}{0.26}         &\meanstd{66.79}{0.66}     \\
                            & LUCIR~\cite{hou2019learning}     & \meanstd{94.30}{0.79}          & \meanstd{94.99}{0.14}          & \meanstd{85.13}{0.20}          & \meanstd{87.50}{0.44}          & \meanstd{70.09}{0.40}          \\  
                            & BiC~\cite{wu2019large}          & \meanstd{95.41}{0.73}          & \textbf{\meanstd{96.45}{0.34}}          & \textbf{\meanstd{85.16}{0.36}}          & \meanstd{87.03}{0.30}          & \meanstd{68.44}{4.40} \\
                            & ER-ACE~\cite{caccia2022erace}    & \meanstd{92.12}{0.62}          & \meanstd{94.09}{0.27}          & \meanstd{77.00}{0.80}          & \meanstd{83.30}{0.54}          & \meanstd{68.91}{0.38}          \\
                            \cline{2-7} 
                            & ER~\cite{riemer2019learning}        & \meanstd{92.06}{0.65}          & \meanstd{93.60}{0.76}          & \meanstd{72.42}{1.74}          & \meanstd{81.34}{1.06}          & \meanstd{64.96}{0.45}          \\
                            & ER w/ BFP (Ours)                 & \meanstd{95.50}{0.41}  \boost{3.44}        & \meanstd{96.11}{0.27} \boost{2.51}         & \meanstd{79.79}{1.67}  \boost{7.38}        & \meanstd{84.16}{1.18} \boost{2.81} & \meanstd{71.43}{0.58}  \boost{6.47}        \\ \cline{2-7} 
                            & DER++~\cite{buzzega2020dark}     & \meanstd{91.71}{0.83}          & \meanstd{93.76}{0.27}          & \meanstd{76.96}{0.24}          & \meanstd{83.59}{0.41}          & \meanstd{71.14}{0.53}          \\
                            & DER++ w/ BFP (Ours)              & \textbf{\meanstd{95.95}{0.22}} \boost{4.23} & \meanstd{96.29}{0.26}  \boost{2.53}        & \meanstd{83.64}{0.64} \boost{6.68}         & \textbf{\meanstd{87.20}{0.32}}  \boost{3.61}        & \textbf{\meanstd{73.07}{0.28}} \boost{1.94} \\
\end{tabular}
\caption{Final Average Accuracies (FAA, in \%) in Class-IL and Task-IL setting of baselines and our methods on various datasets and buffer sizes. The green numbers in parentheses show the absolute improvements brought by BFP over the corresponding ER or DER++ baselines. The proposed BFP method can boost the Class-IL performance by up to 6-8\% in some challenge settings with small buffer sizes. DER++ w/ BFP (Ours) outperforms all baselines in terms of Class-IL accuracies and has Task-IL accuracies that are better or close to the top-performing methods. Mean and standard deviation are computed over 5 runs with different seeds. Joint Training (JT) shows the upper bound performance, where the model is trained on data from all tasks and Finetune (FT) is the lower bound, where the model is trained sequentially without handling forgetting. }
\label{tbl:main_results}
\vspace{-1.5em}
\end{table*}

\subsection{Experimental Setting}
\label{sec:exp-setting}

{\bf Continual Learning Settings}. 
We follow \cite{chaudhry2019continual, wu2019large, buzzega2020dark} and test all methods using both the Class-IL and Task-IL settings in our CL experiments. 
Both Class-IL and Task-IL split the dataset into a sequence of tasks, each containing a disjoint set of classes, while task identifiers are available during testing under Task-IL. 
Task-IL thus has extra advantages during inference (e.g. select proper prediction head) and becomes an easier CL scenario. 
Our work is designed for Class-IL and its Task-IL performance is obtained by only considering the logits within the ground truth task. 

{\bf Datasets}.
We evaluate baselines and our methods on the following datasets using varying buffer sizes:
{\bf Split CIFAR-10} divides the original CIFAR-10~\cite{krizhevsky2009learning} dataset into 5 tasks, with each task composed of 2 classes. Each class includes 5000 training and 1000 testing images of shape 32$\times$32.
{\bf Split CIFAR-100} divides CIFAR-100~\cite{krizhevsky2009learning} into 10 tasks, with 10 classes per task. Each class has 500 training and 100 testing images of shape 32$\times$32. 
{\bf Split TinyImageNet} splits TinyImageNet~\cite{tinyIM} into 10 tasks, with 20 classes per task. Each class contains 500 training images, 50 validation images, and 50 testing images. 
These datasets are challenging and state-of-the-art continual learning methods still fall far behind the Joint Training (JT) baseline, especially in the Class-IL setting as shown in Table~\ref{tbl:main_results}. 

{\bf Metrics}. Following~\cite{buzzega2020dark, masana2020class, boschini2022class, arani2022learning}, we report the performance of each compared method using Final Average Accuracy (FAA). Suppose $a_i^t$ is the testing classification accuracy on the $i^{\text{th}}$ task when the training finishes on task $t$, FAA is the accuracy of the final model averaged all tasks:
\begin{align}
    FAA = \frac{1}{T}\sum_{i=1}^{T} a_i^T. 
\end{align}

We also report the Final Forgetting (FF), which reflects the accuracy drop between the peak performance on one task and its final performance:
\begin{align}
    FF = \frac{1}{T-1}\sum_{i=1}^{T-1} \max_{j \in {1, \cdots, T-1}} (a_i^j - a_i^{T}).
\end{align}
Lower FF means less forgetting and better CL performance. 

{\bf Training details}.
We use ResNet-18~\cite{he2016deep} as the network backbone, and instead of the simple reservoir buffer used in~\cite{buzzega2020dark}, we use class-balanced reservoir sampling~\cite{buzzega2021rethinking} for pushing examples into the buffer. 
All the baselines we compare are updated with this change.
We use an SGD optimizer to optimize the model $f_\theta$ and another SGD+Momentum optimizer with a learning rate $0.1$ for the projection matrix $A$. 
The optimizers and the matrix $A$ are re-initialized at the beginning of each task. 
The network is trained for 50 epochs per task for Split CIFAR-10 and Split CIFAR-100 and 100 epochs per task for Split TinyImageNet. The learning rate is divided by 10 after a certain number of epochs within each task ($[35, 45]$ for Split CIFAR-100 and $[35, 60, 75]$ for Split TinyImageNet). 
In this work, we focus on this offline CL setting where each task is trained for multiple epochs. Although we are also interested in online CL, multi-epoch training helps disentangle underfitting and catastrophic forgetting~\cite{buzzega2020dark, arani2022learning}. 
BFP introduces only one extra hyperparameter $\gamma$, which is set to 1 for all experiments. We found that $\gamma=1$ works well for all datasets and buffer sizes and did not perform hyperparameter searches for individual experiment settings. Hyperparameters $\alpha$ and $\beta$ used in Equation~\ref{eq:total-loss} are adopted from~\cite{buzzega2020dark}.
Most baselines adopt different hyperparameters for different settings, for which we adopt the hyperparameters that have been optimized by grid search by~\cite{buzzega2020dark} and \cite{boschini2022class} for a fair comparison. The details can be found in the~\appendixnote. 


\subsection{Baselines}
\label{sec:baselines}


First, we evaluate the performance of \textbf{Joint Training (JT)} and \textbf{Finetuning (FT)} baselines on each dataset. JT trains the network on all training data, does not have the forgetting problem, and thus indicates the upper-bound performance of CL methods. On the contrary, FT simply performs SGD using the current task data without handling forgetting at all and indicates a lower-bound performance. 

As a feature distillation method, our method can be combined with most continual learning methods. In our evaluation, we test our method by combing it with two popular experience replay methods, ER~\cite{riemer2019learning} and DER++~\cite{buzzega2020dark}. \textbf{ER} uses a memory buffer to store the training examples from past tasks and interleaves them with the current task data for training. 
In addition to this, \textbf{DER++} records the output logits of the examples in the memory and performs logit distillation when doing experience replay. We combine the proposed BFP loss with ER and DER++ and denote them as \textbf{ER w/ BFP} and \textbf{DER++ w/ BFP} respectively. 

We also compare the proposed method with some other state-of-the-art CL baselines as listed in Table~\ref{tbl:main_results}. 
{\bf Incremental Classifier and Presentation Learning (iCaRL)}~\cite{rebuffi2017icarl} performs classification using the nearest mean-of-exemplars, where the exemplars selected by herding algorithm in the feature space. 
{\bf Functional Distance Regularization (FDR)}~\cite{benjamin2018fdr} regularize the output of the network to its past value. Different from DER/DER++, FDR applies the regularization on the output classification probability. 
{\bf Learning a Unified Classifier Incrementally via Rebalancing (LUCIR)}~\cite{hou2019learning} augments experience replay with multiple modifications to preserve old knowledge and enforce separation class separation in continual learning. 
{\bf Bias Correction (BiC)}~\cite{wu2019large} augments the experience replay by learning a separate layer to correct the bias in the output logits. 
{\bf ER with Asymmetric Cross-Entropy (ER-ACE)}~\cite{caccia2022erace} proposes to reduce representation drift by using separate cross-entropy loss for online and replayed training data. 

\subsection{Results}

The Final Average Accuracies in the Class-IL and Task-IL settings are reported in Table~\ref{tbl:main_results}. The corresponding table for Final Forgetting can be found in the~\appendixnote. 
We test the methods on three datasets with various sizes of memory buffers. 
Experiments are averaged over 5 runs with different seeds and mean and standard deviation are reported. 
First, we observe that there is a still big gap between the current best CL methods and the JT oracle baselines on all datasets, especially in the Class-IL setting, which indicates that CL is still an unsolved and challenging problem. 
Comparing DER++ and DER++ w/ BFP, we can see that BFP boosts the Class-IL accuracies by a significant margin, especially with a small buffer size (6.8\% on S-CIFAR10 with buffer size 200 and 8.5\% on S-CIFAR100 with buffer size 500). 
DER++ w/ BFP thus outperforms all baseline methods in the Class-IL setting, which are very challenging as the final model needs to distinguish testing examples from all seen classes. Previous CL methods struggle to have satisfactory performance in this setting. 
Under the Task-IL setting that is easier because task identifiers are known during evaluation time, our model also helps achieve much higher accuracies over the base ER or DER++ method. And among all the CL methods compared, the proposed method also achieves the best or close-to-best accuracies under the Task-IL setting. 


\subsection{Linear Probing}
\label{sec:linear-prob}

\begin{figure}[t!]
    \centering
    \includegraphics[width=\linewidth]{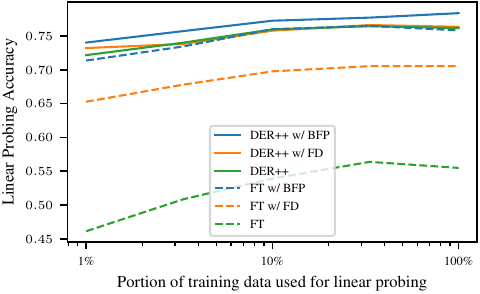}
    \caption{Linear probing accuracies on the frozen feature extractor that is obtained after training on Split-CIFAR10 with 200 buffer size, using different methods. A higher linear probing accuracy indicates a better feature space where data points are more linearly separable. Note that with the help of BFP, even a simple FT baseline can learn a representation as good as the powerful DER++ method. }
    \label{fig:linear-prob}
    \vspace{-1em}
\end{figure}

Some latest work on continual learning studied catastrophic forgetting in the final feature space $h^T(x)$~\cite{davari2022probing, zhang2022feature}. 
They show that although the accuracy using the continually trained classifier $g^T$ degrades heavily due to catastrophic forgetting, the learned knowledge in $h^T$ is well maintained. 
This is shown by fitting a linear classifier $g^*$ on top of the frozen feature extractor at the end of continual learning $h^T$. 
Such \textit{linear probing accuracies} obtained by $g^*\circ h^T$ can be much higher than $g^T \circ h^T$. Therefore recent work argues that catastrophic forgetting mainly happens at the last linear classification layer and the linear probing accuracies can be used as a quality measure for the representation learned from continual learning~\cite{fini2022self}. 
We conduct a similar linear probing analysis on baselines combined with the BFP, and we additionally test the effect of our method on the naive FT baseline, which is denoted as \textbf{FT w/ BFP} in Figure~\ref{fig:linear-prob}. 
In FT w/ BFP, we do not use the memory buffer and thus $\alpha=\beta=0$ in Equation~\ref{eq:bfp}, but we apply the BFP loss on the online data stream with $\gamma=1$. 
Linear probing accuracies on Split CIFAR-10 are reported in Figure~\ref{fig:linear-prob}, where we also vary the portion of training data used for linear probing. 
The results show that BFP boosts the linear probing accuracies of the FT baseline by a significant margin, achieving a similar performance with the powerful experience replay method, DER++. When combined with DER++, BFP also helps improve the linear probing accuracies. 
This indicates that either with or without a memory buffer, BFP helps learn a better feature space during CL, where examples from different classes remain linearly separable.

\subsection{Ablation Study}
\label{sec-abstudy}

\begin{table}[t]
    \footnotesize
    \centering
    \begin{tabular}{l|lccc}
    \hline
    Dataset                     & Buffer & \multicolumn{1}{l}{FD} & \multicolumn{1}{l}{BFP} & \multicolumn{1}{l}{BFP-2} \\ \hline
    \multirow{2}{*}{S-CIFAR10}   & 200    & 68.44±1.35           & \textbf{72.21±0.22}            & 72.04±0.96              \\
                                & 500    & 74.50±0.41           & 76.02±0.79            & \textbf{76.63±0.63}              \\ \hline
    \multirow{2}{*}{S-CIFAR100} & 500    & 43.81±1.35           & \textbf{47.45±1.30}            & \textbf{47.45±1.08}              \\
                                & 2000   & 56.56±0.55           & \textbf{57.91±0.66}            & 57.27±0.67              \\ \hline
    S-TinyImg                   & 4000   & 42.40±1.02           & \textbf{43.40±0.41}            & 42.91±0.50              \\ \hline
    \end{tabular}
    \caption{Class-IL Final Average Accuracy using different types of layers for backward feature projection. The baseline method is DER++. }
    \label{tab:ab-layer}
    \vspace{-1em}
\end{table}

We study the effect of different types of projection layers used for backward feature projection, in Equation~\ref{eq:bfp}. 
Our main results are obtained using a linear projection layer as the learnable transformation layer (denoted as \textbf{BFP}). We also test our method using an identity function $A=I$ as the projector, which is essentially a feature distillation loss (\textbf{FD}) on the final feature, as well as using a non-linear function (a two-layer MLP with ReLU activation in between) as $p$, which is denoted as \textbf{BFP-2}. 
These variations are tested when integrated with DER++~\cite{buzzega2020dark} and the results are shown in Table~\ref{tab:ab-layer}. 
According to Table~\ref{tab:ab-layer}, while the simple FD method already outperforms the baseline, the proposed learnable BFP further boosts the accuracies by a large margin. 
This is expected because FD regularizes the learned features directly to those from the old model, while the old model has not learned from the new data and may give useless features. 
In this case, FD promotes stability while lacking plasticity. On the contrary, BFP is learnable and thus provides the desired plasticity that allows new knowledge to appear while maintaining the old ones. 
Furthermore, we can also see that the performance already saturates with a linear projection layer and a more complex non-linear projection (BFP-2) does not improve further. We hypothesize that because BFP is applied on the feature space just before the linear classifier, linear separability is better maintained with a linear transformation rather than a non-linear function. 

\subsection{Feature Similarity Analysis}
\label{sec:feat-sim}

\begin{figure}
    \centering
    \includegraphics[width=\linewidth]{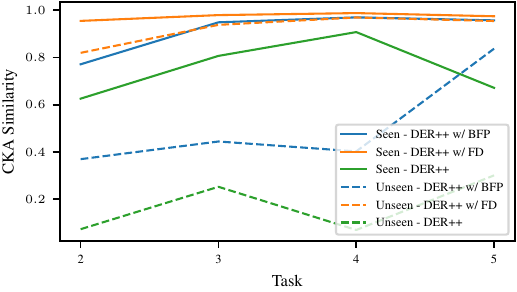}
    \caption{Feature similarity at different tasks of training on Split-CIFAR10, with 200 buffer size, using different CL methods.}
    \label{fig:cka}
    \vspace{-1em}
\end{figure}

To demonstrate that the proposed BFP method regularizes the features that are already learned while allowing features of new data to freely evolve, we conduct an analysis of feature similarity. 
Following~\cite{ramasesh2020anatomy, davari2022probing}, we adopt Centered Kernal Alignment (CKA)~\cite{kornblith2019similarity} to measure the feature similarity before and after training on a task. CKA is a similarity measure for deep learned representations, and it's invariant to isotropic scaling and orthogonal transformation~\cite{kornblith2019similarity}.
CKA between two feature matrices $Z_1\in \RR^{d_1\times n}$ and $Z_2\in \RR^{d_2\times n}$ with a linear kernel is defined as 
\begin{align}
    \cka(Z_1, Z_2) = \frac{\| Z_1 Z_2^T \|_F^2}{\|Z_1 Z_1^T\|_F^2 \|Z_2 Z_2^T\|_F^2}.
\end{align}
Recall that the feature matrix extracted from $\Data_i$ using model $h^j$ is denoted as $Z_i^j = h^j(\Data_i)$, and similar $Z_{1:i}^j = h^i(\Data_{1:i})$. 
During learning on task $t$, we consider two sets of features, features from $\Data_{1:t-1}$ that have been learned by the model (seen) and features from $\Data_{t}$ that are new to the model (unseen). We define their CKA similarity before and after learning on task $t$ respectively as follows
\begin{align}
    \cka_{t}^\text{seen} =& \; \cka(Z_{1:t-1}^{t-1}, Z_{1:t-1}^{t}) \\
    \cka_{t}^\text{unseen} =& \; \cka(Z_{t}^{t-1}, Z_{t}^{t}). 
\end{align}

Note that $Z_{t}^{t-1}$ represents the features extracted from future data by the old model, and it's expected that they do not provide useful information. On the contrary, $Z_{1:t-1}^{t-1}$ has already been well learned and we want to preserve its structure. 
Therefore, we want $\cka_{t}^\text{seen}$ to be high to retain knowledge, while $\cka_{t}^\text{unseen}$ low to allow the feature of unseen data to change freely in continual learning. 
We plot $\cka_{t}^\text{seen}$ and $\cka_{t}^\text{unseen}$ during CL in Figure~\ref{fig:cka} and the results confirm our desire. 
DER++ applies no direct constraint on the feature space during CL and thus similarity is low for both seen and unseen data. 
On the contrary, FD poses a strong constraint on both seen and unseen data, resulting in high similarities. In this way, FD gains more stability at the cost of lower plasticity. 
Combining their respective advantages, BFP keeps a high $\cka_{t}^\text{seen}$ while allowing the unseen features to change (low $\cka_{t}^\text{unseen}$), and thus is able to achieve a better trade-off between stability and plasticity.


\section{Conclusion}
\label{sec:conclusion}

In this paper, we reduce catastrophic forgetting in continual learning (CL) by proposing Backward Feature Projection (BFP), a learnable feature distillation method. 
We show that during CL, despite the large dimension of the feature space, only a small number of feature directions are used for classification. Without regularization, previously learned feature directions diminish and harm linear separability, resulting in catastrophic forgetting. 
The proposed BFP helps maintain linear separability learned from old tasks while allowing new feature directions to be learned for new tasks. In this way, BFP achieves a better trade-off between plasticity and stability. 
BFP can be combined with existing experience replay methods and experiments show that it can boost performance by a significant margin. 
We also show that BFP results in a more linearly separable feature space, on which a linear classifier can recover higher accuracies.

{\small
\bibliographystyle{ieee_fullname}
\bibliography{main}
}

\newpage
\appendix
\section{\crchange{Acknowledgements}}

The authors thank Jongseong Jang, Yizhou (Philip) Huang, Kevin Xie, and Nikita Dhawan for discussions and useful feedback. 

\section{Implementation Details}

\subsection{Complete Algorithm}

In Algorithm~\ref{alg:main}, we provide the pseudocode of continual learning with the proposed BFP method. Note that following~\cite{buzzega2020dark}, we sample training data points from the memory buffer for each loss independently. 
We empirically find this results in better performance than using the same set of replayed samples for all losses. 
The images without augmentation $x_o$ are pushed into the memory and replayed images are augmented on the fly. 
The classification model is trained using an SGD optimizer (\textit{sgd}) and the projection matrix $A$ is trained using an SGD+Momentum optimizer (\textit{sgdm}). 

\begin{algorithm}[b]
\small
\caption{- Continual Learning with BFP}
\label{alg:main}
\begin{algorithmic}
  \STATE {\bfseries Input:} dataset $\{\Data_1, \cdots, \Data_T\}$, parameters $\theta=\{\phi, \psi\}$, scalars $\alpha$, $\beta$ and $\gamma$, optimizer $\text{\textit{sgd}}, \text{\textit{sgdm}}$,
  \STATE $M \gets \{\}$
  \FOR {\texttt{$t$} \textbf{from} \texttt{$1$} \textbf{to} \texttt{$T$}}
      \STATE $A \gets \text{\textit{random-init}}()$
      \STATE $\text{\textit{sgdm}} \gets \text{\textit{reinit}}(\text{\textit{sgdm}})$
      \FOR{\texttt{$(x_{o},y_{o})$} \textbf{in}
          \texttt{$\Data_t$}}
          \STATE $x, y \gets \text{\textit{augment}}(x_o, y_o)$
          \STATE $L \gets \text{\textit{cross-entropy}} (y, f_\theta(x))$ \hfill\COMMENT{Eq.~\ref{eq:loss-ce}}
          
          \IF{$t>1$}
          \STATE $x, y \gets \text{\textit{augment}}(\text{\textit{sample}}(M))$
          \STATE $L_\text{rep-ce} \gets \text{\textit{cross-entropy}} (y, f_\theta(x))$ \hfill\COMMENT{Eq.~\ref{eq:loss-repce}}
          \STATE $x, y \gets \text{\textit{augment}}(\text{\textit{sample}}(M))$
          \STATE $L_\text{rep-logits} \gets \|f_\theta (x) - f_{\text{old}}(x)\|_2$  \hfill\COMMENT{Eq.~\ref{eq:loss-replogits}}
          \STATE $x, y \gets \text{\textit{augment}}(\text{\textit{sample}}(M))$
          \STATE $L_\text{BFP} \gets \| Ah_\psi(x) - h_\text{old}(x) \|_2$  \hfill\COMMENT{Eq.~\ref{eq:loss-BFP}}
          \STATE $L = L + L_\text{rep-ce} + L_\text{rep-logits} + L_\text{BFP}$  \hfill\COMMENT{Eq.~\ref{eq:total-loss}}
          \ENDIF
          
          \STATE $\theta \gets \text{\textit{sgd}} (\theta, \nabla_\theta L)$
          \STATE $A \gets \text{\textit{sgdm}} (A, \nabla_A L)$
          
          \STATE $M \gets \text{\textit{balanced-reservoir}}(M, (x_o, y_o))$ \hfill\COMMENT{Alg.~\ref{alg:balancoir}}
      \ENDFOR
      \STATE $f_\text{old} = \text{\textit{freeze}} (f_\theta)$
    \ENDFOR
\end{algorithmic}
\end{algorithm}

\subsection{Class-balanced Reservoir Sampling}

We adopt the class-balanced reservoir sampling (BRS)~\cite{buzzega2021rethinking} for memory buffer management. The detail of this algorithm is described in Algorithm~\ref{alg:balancoir}. Compared to regular Reservoir Sampling (RS), BRS ensures that every class has an equal number of examples stored in the memory buffer. All experiments are incorporated with this change. Empirically we find that BRS does not bring significant changes compared to RS, but it helps to reduce variance in the results. 

\begin{algorithm}[t]
    \caption{Balanced Reservoir Sampling~\cite{buzzega2021rethinking}}
    \label{alg:balancoir}
    \begin{algorithmic}[1]
      \STATE {\bfseries Input:} replay buffer $M$, exemplar $(x,y)$, 
      \STATE \hspace{2.70em} number of seen examples $N$.
      \IF{$|M| > N$}
            \STATE $M[N]\gets (x,y)$
      \ELSE
            \STATE $j \gets \operatornamewithlimits{RandInt}([0, N])$
            \IF {$j < |M|$}
                \vspace{5.5mm}
                \STATE \tikzmark{start6}$M[j]\gets (x,y)$  \tikzmark{end6} \\ \vspace{-9.4mm} \hspace{-0.7em} \textbf{Reservoir Sampling} 
                \vspace{3.2em}
                \vspace{1mm}
                \STATE $\tilde{y} \gets \small{\argmax} \ \operatornamewithlimits{ClassCounts}(M,y)$
                \STATE\tikzmark{start7}$k\gets  \operatornamewithlimits{RandChoice}(\{\tilde{k}; M[\tilde{k}] = (x,y), y = \tilde{y}\})$ \tikzmark{end7} 
                \STATE $M[k]\gets (x,y)$ \\ \vspace{-5em} \hspace{-0.7em} \textbf{Balanced Reservoir Sampling} \vspace{4em}
                \vspace{1.2mm}
            \ENDIF
      \ENDIF
      \vspace{-1.2em}
    \end{algorithmic}
\TextboxReservoir[0]{start6}{end6}{}
\TextboxBalancedReservoir[0]{start7}{end7}{}
\end{algorithm}


\subsection{Training details}

Image sizes are $32\times32$ in Split-CIFAR10 and Split-CIFAR100 and $64\times64$ in Split-TinyImageNet. 
All experiments use the same data augmentation procedure, applied on input from both the current task and the memory buffer independently. Data augmentation includes a full-size random crop with a padding of 4 pixels, a random horizontal flip, and normalization. 

For all experiments involving BFP, the optimizer for the matrix $A$ is an SGD+Momentum optimizer with a learning rate of 0.1 and momentum of 0.9. The weighting term $\gamma$ in Equation~\ref{eq:total-loss} is 1. Empirically we find that the BFP performance is not sensitive to these hyperparameters, and we use this one set of hyperparameters for BFP loss in all experiments. 

\subsection{Hyperparameters}

In this section, we list the best hyperparameters used for the compared baselines mentioned in Section~\ref{sec:baselines} and their results are reported in Table~\ref{tbl:main_results}. 
These hyperparameters are adopted from~\cite{buzzega2020dark} and~\cite{boschini2022class}, where they were selected by a hyperparameter search conducted on a held-out 10\% training set for validation. Please refer to~\cite{buzzega2020dark, boschini2022class} for further details.

\crchange{The proposed BFP only introduced a single hyperparameter $\gamma$, which is set to a constant value of $1$ throughout all experiments and does not need extra tuning. Other hyperparameters like $\alpha$ and $\beta$ are inherited from ER and DER++~\cite{buzzega2020dark} and we simply adopt} the same set of hyperparameters from~\cite{buzzega2020dark}. We do not further tune or modify them. 


\subsubsection{Split CIFAR-10}

\noindent{\bf FT}: $lr=0.1$

\noindent{\bf JT}: $lr=0.1$

\vspace{0.5em}
\noindent{\bf Buffer size = 200}

\noindent\fbox{%
    \parbox{\linewidth}{%
\noindent{\bf iCaRL}: $lr=0.1$, $wd=10^{-5}$

\noindent{\bf FDR}: $lr=0.03$, $\alpha=0.3$

\noindent{\bf LUCIR}: $\lambda_{\textrm{base}}=5$, $mom=0.9$, $k=2$, $\text{epoch}_{\text{fitting}}=20$, $lr=0.03$, $\text{lr}_\text{fitting}=0.01$, $m=0.5$

\noindent{\bf BiC}: $\tau=2$, $\text{epochs}_\text{BiC}=250$, $lr=0.03$

\noindent{\bf ER-ACE}: $lr=0.03$

\noindent{\bf ER}: $lr=0.1$

\noindent{\bf DER++}: $lr=0.03$, $\alpha=0.1$, $\beta=0.5$
    }%
}

\vspace{0.5em}
\noindent{\bf Buffer size = 500}

\noindent\fbox{%
    \parbox{\linewidth}{%
\noindent{\bf iCaRL}: $lr=0.1$, $wd=10^{-5}$

\noindent{\bf FDR}: $lr=0.03$, $\alpha=1$

\noindent{\bf LUCIR}: $\lambda_{\textrm{base}}=5$, $mom=0.9$, $k=2$, $\text{epoch}_{\text{fitting}}=20$, $lr=0.03$, $\text{lr}_\text{fitting}=0.01$, $m=0.5$

\noindent{\bf BiC}: $\tau=2$, $\text{epochs}_\text{BiC}=250$, $lr=0.03$

\noindent{\bf ER-ACE}: $lr=0.03$

\noindent{\bf ER}: $lr=0.1$

\noindent{\bf DER++}: $lr=0.03$, $\alpha=0.2$, $\beta=0.5$
    }%
}

\subsubsection{Split CIFAR-100}

\noindent{\bf FT}: $lr=0.03$

\noindent{\bf JT}: $lr=0.03$

\vspace{0.5em}
\noindent{\bf Buffer size = 500}

\noindent\fbox{%
    \parbox{\linewidth}{%
\noindent{\bf iCaRL}: $lr=0.3$, $wd=10^{-5}$

\noindent{\bf FDR}: $lr=0.03$, $\alpha=0.3$

\noindent{\bf LUCIR}: $\lambda_{\textrm{base}}=5$, $mom=0.9$, $k=2$, $\text{epoch}_{\text{fitting}}=20$, $lr=0.03$, $\text{lr}_\text{fitting}=0.01$, $m=0.5$

\noindent{\bf BiC}: $\tau=2$, $\text{epochs}_\text{BiC}=250$, $lr=0.03$

\noindent{\bf ER-ACE}: $lr=0.03$

\noindent{\bf ER}: $lr=0.1$

\noindent{\bf DER++}: $lr=0.03$, $\alpha=0.2$, $\beta=0.5$
    }%
}

\vspace{0.5em}
\noindent{\bf Buffer size = 2000}

\noindent\fbox{%
    \parbox{\linewidth}{%
\noindent{\bf iCaRL}: $lr=0.3$, $wd=10^{-5}$

\noindent{\bf FDR}: $lr=0.03$, $\alpha=1$

\noindent{\bf LUCIR}: $\lambda_{\textrm{base}}=5$, $mom=0.9$, $k=2$, $\text{epoch}_{\text{fitting}}=20$, $lr=0.03$, $\text{lr}_\text{fitting}=0.01$, $m=0.5$

\noindent{\bf BiC}: $\tau=2$, $\text{epochs}_\text{BiC}=250$, $lr=0.03$

\noindent{\bf ER-ACE}: $lr=0.03$

\noindent{\bf ER}: $lr=0.1$

\noindent{\bf DER++}: $lr=0.03$, $\alpha=0.1$, $\beta=0.5$
    }%
}

\subsubsection{Split TinyImageNet}

\noindent{\bf FT}: $lr=0.03$

\noindent{\bf JT}: $lr=0.03$

\vspace{0.5em}
\noindent{\bf Buffer size = 4000}

\noindent\fbox{%
    \parbox{\linewidth}{%
\noindent{\bf iCaRL}: $lr=0.03$, $wd=10^{-5}$

\noindent{\bf FDR}: $lr=0.03$, $\alpha=0.3$

\noindent{\bf LUCIR}: $\lambda_{\textrm{base}}$=5, $mom$=0.9, $k$=2, $\text{epoch}_{\text{fitting}}$=20, $lr$=0.03, $\text{lr}_\text{fitting}$=0.01, $m$=0.5

\noindent{\bf BiC}: $\tau=2$, $\text{epochs}_\text{BiC}=250$, $lr=0.03$

\noindent{\bf ER-ACE}: $lr=0.03$

\noindent{\bf ER}: $lr=0.1$

\noindent{\bf DER++}: $lr=0.1$, $\alpha=0.3$, $\beta=0.8$
    }%
}

\section{Additional results}

\subsection{Final Forgetting}

Final Forgetting (FF) measures the performance drop between the end of each task and the end of CL. A CL method with a lower FF has a better ability to retain knowledge during CL and thus better stability. However, higher stability may come with the price of plasticity, and we remind readers that the Final Average Accuracy (FAA) reported in Table~\ref{tbl:main_results} can better reflect the trade-off between stability and plasticity. 
The Final Forgetting for all baselines and our methods can be found in Table~\ref{tbl:main_forgetting}. As we can see from Table~\ref{tbl:main_forgetting}, in the class-IL setting, the proposed DER++ w/ BFP method helps reduce FF compared to the base DER++ method by 11\% and 12\% on S-CIFAR10 with 200 buffer size and S-CIFAR100 with 500 buffer size respectively. DER++ w/ BFP also achieves the lowest FF among all compared methods in the class-IL setting. 
Final Forgettings in the Task-IL setting are generally much lower than those from the Class-IL setting because Task-IL provides the oracle task identifiers during the testing time and thus becomes a much easier CL scenario. In this setting, the proposed BFP also brings large improvements over the base ER and DER++ methods. 

\begin{table}[th]
    \footnotesize
    \centering
    \begin{tabular}{l|lccc}
    \hline
    Dataset                     & Buffer & \multicolumn{1}{l}{FD} & \multicolumn{1}{l}{BFP} & \multicolumn{1}{l}{BFP-2} \\ \hline
    \multirow{2}{*}{S-CIFAR10}   & 200    & 55.10±1.85 &  \textbf{63.27±1.09} &  60.61±2.72             \\
                                & 500    & 66.37±1.37 &  \textbf{71.51±1.58} &  70.25±1.18             \\ \hline
    \multirow{2}{*}{S-CIFAR100} & 500    & 20.02±0.09 &  \textbf{22.54±1.10} &  21.25±0.73            \\
                                & 2000   & 36.81±0.71 &  38.92±1.94 &  \textbf{39.42±2.54}           \\ \hline
    S-TinyImg                   & 4000   & 23.13±0.77 &  \textbf{26.33±0.68} &  25.87±0.86             \\ \hline
    \end{tabular}
    \caption{Class-IL Final Average Accuracy using different types of layers for backward feature projection. The base method is ER. }
    \label{tab:ab-layer-er}
    \vspace{-1em}
\end{table}

\begin{table*}[t]
\centering
\footnotesize
\begin{tabular}{ll|lllll}
\multicolumn{1}{c}{Setting} & \multicolumn{1}{c|}{Method}      & \multicolumn{2}{c}{S-CIFAR10}                                                                         & \multicolumn{2}{c}{S-CIFAR100}                                                                        & S-TinyImageNet                                    \\ \hline \hline
\multicolumn{1}{r}{}        & \multicolumn{1}{c|}{Buffer Size} & 200                                               & 500                                               & 500                                               & 2000                                              & 4000                                              \\ \hline \hline
Class-IL                    & Joint Training                               & \multicolumn{2}{c}{-}                                          & \multicolumn{2}{c}{-}                                          & -          \\
                            & Finetune                               & \multicolumn{2}{c}{\meanstd{96.44}{0.28}}                                          & \multicolumn{2}{c}{\meanstd{89.54}{0.16}}                                           & \meanstd{78.54}{0.45}           \\ 
                            & iCaRL~\cite{rebuffi2017icarl}   & \meanstd{27.75}{0.82}&\meanstd{25.31}{4.35}&\meanstd{30.13}{0.28}&\meanstd{24.72}{0.66}&\meanstd{16.82}{0.51}         \\
                            & FDR~\cite{benjamin2018fdr}      & \meanstd{76.08}{4.87}&\meanstd{83.16}{4.72}&\meanstd{73.71}{0.68}&\meanstd{60.90}{1.41}&\meanstd{57.01}{0.59} \\
                            & LUCIR~\cite{hou2019learning}     & \meanstd{46.36}{3.17}&\meanstd{29.11}{0.63}&\meanstd{53.24}{0.56}&\meanstd{34.16}{1.19}&\meanstd{25.50}{1.86}          \\ 
                            & BiC~\cite{wu2019large}          & \meanstd{44.36}{2.73}&\meanstd{20.88}{2.17}&\meanstd{51.86}{1.57}&\meanstd{41.42}{1.61}&\meanstd{61.67}{1.42}         \\
                            & ER-ACE~\cite{caccia2022erace}   & \meanstd{21.59}{1.19}&\meanstd{15.07}{0.99}&\meanstd{38.32}{1.29}&\meanstd{28.69}{0.87}&\meanstd{30.83}{0.23}        \\
                            \cline{2-7} 
                            & ER~\cite{riemer2019learning}     &\meanstd{42.19}{5.19}&\meanstd{26.64}{6.33}&\meanstd{47.62}{33.70}&\meanstd{44.03}{18.79}&\meanstd{49.61}{16.47}   \\
                            & ER w/ BFP (Ours)                &\meanstd{32.23}{5.58} \reduce{9.96} &\meanstd{22.67}{6.64} \reduce{3.97}&\meanstd{47.69}{30.30} \reduce{0.07}&\meanstd{37.49}{18.06} \reduce{6.54} &\meanstd{41.59}{20.77} \reduce{8.02}       \\ \cline{2-7} 
                            & DER++~\cite{buzzega2020dark}   &\meanstd{28.28}{1.06}&\meanstd{20.16}{1.49}&\meanstd{42.58}{2.03}&\meanstd{26.29}{1.66}&\meanstd{16.03}{1.20}       \\
                            & DER++ w/ BFP (Ours)            &\textbf{\meanstd{16.69}{0.28}} \reduce{11.59} &\textbf{\meanstd{13.25}{0.64}} \reduce{6.91} &\textbf{\meanstd{29.85}{0.97}} \reduce{12.73} &\textbf{\meanstd{20.91}{0.86}} \reduce{5.39} &\textbf{\meanstd{9.42}{1.04}} \reduce{6.28}    \\ \hline \hline
Task-IL                     & Joint Training                               & \multicolumn{2}{c}{-}                                          & \multicolumn{2}{c}{-}                                          & -          \\
                            & Finetune                               & \multicolumn{2}{c}{\meanstd{39.72}{6.27}}                                          & \multicolumn{2}{c}{\meanstd{60.46}{2.74}}                                          & \meanstd{67.04}{1.27}          \\  
                            & iCaRL~\cite{rebuffi2017icarl}    & \meanstd{4.29}{1.00}&\meanstd{1.91}{2.12}&\meanstd{3.67}{0.40}&\meanstd{1.82}{0.32}&\meanstd{3.56}{0.46}      \\
                            & FDR~\cite{benjamin2018fdr}      &\meanstd{7.03}{1.38}&\meanstd{4.47}{0.45}&\meanstd{16.63}{0.20}&\meanstd{9.17}{0.33}&\meanstd{13.73}{0.30}      \\
                            & LUCIR~\cite{hou2019learning}     & \meanstd{2.83}{0.99}&\meanstd{2.04}{0.27}&\textbf{\meanstd{2.61}{0.17}}&\textbf{\meanstd{1.08}{0.13}}&\meanstd{4.95}{0.61}      \\  
                            & BiC~\cite{wu2019large}          &  \textbf{\meanstd{0.81}{0.77}}&\textbf{\meanstd{0.24}{0.25}}&\meanstd{3.95}{0.35}&\meanstd{2.36}{0.40}&\meanstd{7.08}{3.74}      \\
                            & ER-ACE~\cite{caccia2022erace}    &  \meanstd{6.10}{0.72}&\meanstd{3.64}{0.29}&\meanstd{13.95}{0.45}&\meanstd{7.36}{0.43}&\meanstd{10.67}{0.41}   \\
                            \cline{2-7} 
                            & ER~\cite{riemer2019learning}      &\meanstd{5.71}{0.60}&\meanstd{3.54}{1.15}&\meanstd{11.55}{6.31}&\meanstd{6.12}{2.49}&\meanstd{11.77}{2.06}    \\
                            & ER w/ BFP (Ours)                &\meanstd{1.38}{0.29} \reduce{4.34}&\meanstd{0.77}{0.38} \reduce{2.77}&\meanstd{5.63}{1.56} \reduce{5.92}&\meanstd{2.95}{0.75} \reduce{3.16}&\textbf{\meanstd{3.31}{1.19}} \reduce{8.46}  \\ \cline{2-7} 
                            & DER++~\cite{buzzega2020dark}    &\meanstd{3.88}{0.51}&\meanstd{1.65}{0.17}&\meanstd{11.68}{0.55}&\meanstd{4.80}{0.45}&\meanstd{6.73}{0.41}     \\
                            & DER++ w/ BFP (Ours)             &\meanstd{1.04}{0.23} \reduce{2.84}&\meanstd{0.53}{0.23} \reduce{1.12}&\meanstd{6.36}{0.43} \reduce{5.32}&\meanstd{3.26}{0.15} \reduce{1.54}&\meanstd{4.17}{0.37} \reduce{2.49} \\
\end{tabular}
\caption{Final Forgetting (FF, in \%, lower is better) in Class-IL and Task-IL setting of baselines and our methods on various datasets and buffer sizes. The green numbers in parentheses show the absolute improvements over the corresponding ER or DER++ baselines brought by BFP. }
\label{tbl:main_forgetting}
\end{table*}

\subsection{Ablation Study based on Experience Replay}

We conduct the same ablation study as that in Section~\ref{sec-abstudy}, on different types of the projection layer used in ER w/ BFP. The results are reported in Table~\ref{tab:ab-layer-er}. From Table~\ref{tab:ab-layer-er}, we can draw the same conclusion as in Section~\ref{sec-abstudy}. BFP uses learnable linear transformation when distilling features and thus results in better plasticity during CL compared to the simple FD method. Results show that BFP outperforms FD by a significant margin and has better performance than BFP-2 in most cases. This further shows that enforcing a linear relationship between the new and old features could better preserve linear separability and result in less forgetting in CL.

\subsection{Linear Probing}

\begin{figure}[b!]
    \centering
    \includegraphics[width=\linewidth]{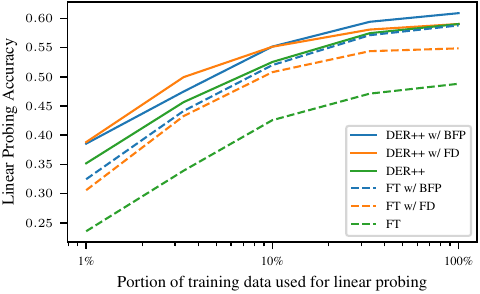}
    \includegraphics[width=\linewidth]{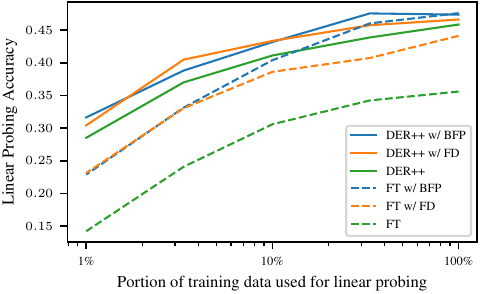}
    \caption{Linear probing accuracies on the fixed feature extractor obtained after training on Split-CIFAR100 (top) and TinyImageNet (bottom). DER++ and its variants use a buffer size of 500 for CIFAR100 and 4000 for TinyImageNet.}
    \label{fig:linear-prob-supp}
\end{figure}

We conduct the same linear probing analysis as Section~\ref{sec:linear-prob} Figure~\ref{fig:linear-prob} on Split-CIFAR100 and Split-TinyImageNet, and the results are reported in Figure~\ref{fig:linear-prob-supp}. On these two datasets, while FD and BFP result in similar linear probing performance when based on DER++, BFP still leads to better linear probing accuracies when based on FT, especially when a large subset of training data is used for linear probing. FT w/ BFP (without the memory buffer) has a similar or even better performance than DER++ (with the memory buffer). This shows that BFP help learns a better feature space from CL, where features from different class are more linearly separable. 

\subsection{Feature Similarity Analysis}

\begin{figure}[t!]
    \centering
    \includegraphics[width=\linewidth]{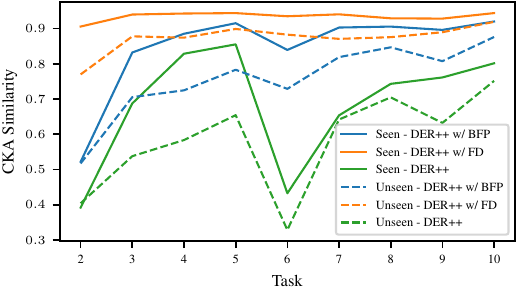}
    \includegraphics[width=\linewidth]{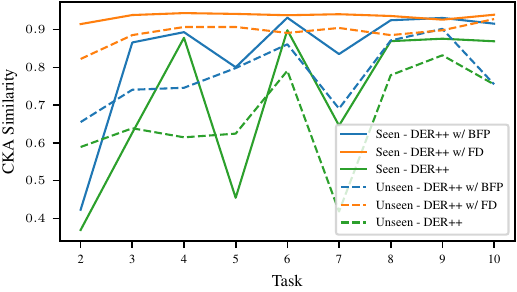}
    \caption{Feature similarity at different tasks of training on Split-CIFAR100 with buffer size 500 (top) and Split-TinyImageNet with buffer size 4000 (bottom), using different CL methods.}
    \label{fig:cka-supp}
\end{figure}

We perform the same feature similarity analysis as Section~\ref{sec:feat-sim} and Figure~\ref{fig:cka} on Split-CIFAR100 and Split-TinyImageNet, and the results are reported in Figure~\ref{fig:cka-supp}. 
From Figure~\ref{fig:cka-supp}, although the curves have high variance throughout continual learning, we can see that BFP has feature similarities that are higher than the DER++ baseline but lower than the naive FD, and thus achieve a better trade-off between stability and plasticity. 

\begin{table}[htb!]
    \small
    \centering
    \begin{tabular}{l|llll}
    \hline
    Method & DER++ & w/ FD & w/ BFP & w/ BFP-2 \\ \hline
    Class-IL   & \meanstd{49.20}{1.99}  & \meanstd{51.89}{3.42} & \meanstd{\textbf{54.45}}{0.86}  & \meanstd{52.88}{1.86}    \\
    Task-IL    & \meanstd{69.01}{2.01}  & \meanstd{71.23}{2.80} & \meanstd{\textbf{72.05}}{1.04}  & \meanstd{70.56}{1.47}     \\ \hline
    \end{tabular}
    \vspace{-1.2em}
    \caption{Final Average Accuracy on ImageNet-100. (mean$\pm$std over 3 runs)}
    \vspace{-1.2em}
    \label{tab:imn100}
\end{table}

\subsection{\crchange{Experiments on Split-ImageNet100}}

To demonstrate that the proposed BFP method scales to large datasets, we conduct experiments on ImageNet100~\cite{rebuffi2017icarl, hou2019learning}. We split ImageNet100 into 10 tasks with 10 classes per task and use a memory buffer of size 2000. The model is trained for 65 epochs on each task using an SGD optimizer with an initial learning rate of 0.1 and weight decay of $5\times 10^{-4}$. Within each task, the learning rate goes through a linear warm-up scheduler for the first 5 epochs and then decays with a 0.1 rate after 15, 30, 40, and 45 epochs. 
The results are reported in Table~\ref{tab:imn100}, which shows that the proposed BFP method still gives a significant improvement (over 5\% in Class-IL setting) over the DER++ baseline, confirming our existing results. 

\subsection{\crchange{Effect of $\gamma$ on Plasticity and Stability}}

In continual learning, the weight of regularization loss controls how closely and strictly the model should resemble the old checkpoints. Therefore the weight serves as a control knob on the trade-off between stability and plasticity: with a stronger regularization loss, the model forgets old tasks less but instead has a hard time learning new tasks.

Although we did not perform an extensive hyperparameter search on $\gamma$ for individual combinations of datasets and buffer sizes, we are still interested in how the varying $\gamma$ affects the trade-off between stability and plasticity in continual learning. Therefore, we train DER++ w/ BFP on S-CIFAR10 with different $\gamma$ and report the performance in Table~\ref{tab:gamma}. 
Besides FAA and FF, we also report the Average Learning Accuracy (ALA)~\cite{riemer2019learning}, which measures the learning ability on new tasks in continual learning and thus reflects the plasticity. Using the notation from Sec.~\ref{sec:exp-setting}, ALA is defined as
\begin{align}
    ALA=\frac{1}{T} \sum_{i=1}^T a_i^i. 
\end{align}
From Table~\ref{tab:gamma}, we can see that the effect of $\gamma$ aligns with our intuition. A higher $\gamma$ poses a strong regularization on the feature space, resulting in a lower FF (more stable) but also a lower ALA (less plastic). Also, we can observe that the final performance (FAA) remains robust to the value of $\gamma$ within a considerable range. 

\section{\crchange{More Related Work}}

There has been some recent work that also employs PCA computation in continual learning. Note that the proposed BFP does not require PCA computation during training and the feature directions are learned implicitly when optimizing matrix $A$. However, to provide a complete understanding of the literature, we briefly review the related work that also uses PCA for continual learning. 

Doan~\etal~~\cite{doan2021theoretical} proposed PCA-OGD, which combines PCA analysis with Orthogonal Gradient Descent (OGD). PCA-OGD projects the gradients onto the residuals subspace to reduce the interference of gradient updates from the new tasks on the old tasks. 
Zhu~\etal~\cite{zhu2021class} decomposed the learned features during CL using PCA. They showed that feature directions with larger eigenvalues have larger similarities (corresponding angles) before and after learning a task. They proposed that these feature directions are more transferable and less forgettable. They showed that their dual augmentation method can encourage learned features to have more directions with large eigenvalues. 
GeoDL~\cite{simon2021geodl} constructs low-dimensional manifolds for the features extracted by the online model and the old checkpoints and performs knowledge distillation on the manifolds. PCA computation is explicitly conducted on the learned features for the manifold construction. 
SPACE~\cite{saha2021space} used PCA analysis for network compression and pruning in continual learning. Similar to our analysis, they use PCA to split the learned filters in a network into Core, which is important for the current task, and Residual, which can be compressed and freed up to learn future tasks. In their work, PCA computation is required during continual learning on every layer of the network in order to do pruning, This poses a significant computational overhead in CL compared to our BFP. 
Instead of applying PCA analysis in continual learning, Zhang~\etal~\cite{zhang2021monitoring} designed a modified PCA algorithm based on EWC~\cite{lee2017overcoming} so that it has continual learning ability. They aim to reduce the forgetting problem in monitoring multimode processes. 

\begin{table}[htb!]
    \centering
    \begin{tabular}{l|lllll}
    \hline
        $\gamma$&0.1&0.3&1.0&3.0&10.0 \\ \hline
        FAA&74.56&75.77&76.68&76.00&73.54 \\
        FF&16.11&14.63&13.16&13.07&12.69 \\
        ALA&87.45&87.32&87.21&86.45&83.70 \\
    \hline
    \end{tabular}
    \caption{Results on CIFAR10 (buffer size 500) with different $\gamma$. }
    \label{tab:gamma}
\end{table}

\end{document}